\documentclass[letterpaper, 10 pt, conference]{ieeeconf}  

\IEEEoverridecommandlockouts                              

\usepackage{xcolor} 
\usepackage{amsfonts} 
\usepackage{amsmath}
\usepackage{amssymb}
\usepackage{mathtools}

\usepackage{tikz}
\usetikzlibrary{decorations.pathreplacing,calc}
\newcommand{\tikzmark}[1]{\tikz[overlay,remember picture] \node (#1) {};}

\def\HiLi{\leavevmode\rlap{\hbox to \hsize{\color{yellow!50}\leaders\hrule height .8\baselineskip depth .5ex\hfill}}}

\newcommand*{\AddNote}[4]{%
    \begin{tikzpicture}[overlay, remember picture]
        \draw [decoration={brace,amplitude=0.5em},decorate,ultra thin,black]
            ($(#3)!(#1.north)!($(#3)-(0,1)$)$) --  
            ($(#3)!(#2.south)!($(#3)-(0,1)$)$)
                node [align=center, text width=.5cm, pos=0.5, anchor=west] {#4};
    \end{tikzpicture}
}%

\usepackage{mathtools}
\DeclarePairedDelimiter\ceil{\lceil}{\rceil}

\newcommand{\R}{\mathbb{R}}
\newcommand{\X}{\mathcal{X}}
\newcommand{\adds}{\overset{\mathrm{+}}{\gets}}

\newcommand{\collisionPenalty}{\mathcal{CP}}

\newcommand{\edgeQueue}{\mathcal{Q}_E}
\newcommand{\vertexQueue}{\mathcal{Q}_V}
\newcommand{\tree}{\mathcal{T}}
\newcommand{\bestCost}{c_i}
\newcommand{\bestCostBITKOMO}{c_{\text{best}}}
\newcommand{\maxCost}{c_\text{max}}

\newcommand{\unconnectedSet}{X_{\text{uc}}}

\newcommand{\level}{\mathcal{L}}

\usepackage{balance} 

\usepackage{hyperref} 

\usepackage{listings}
\usepackage{algorithm}
\usepackage{algpseudocode}

\usepackage{comment}

\usepackage{graphicx}
\usepackage{subcaption}
\usepackage[font=small,labelfont=bf]{caption}

\definecolor{bitkomogreen}{rgb}{0,0.6,0}

\author{Jay Kamat$^{1,2}$, Joaquim Ortiz-Haro$^{1}$, Marc Toussaint$^{1}$, Florian T. Pokorny$^{3}$, Andreas Orthey$^{1}$%
\thanks{The research has been supported by the Deutsche Forschungsgemeinschaft (DFG, German Research Foundation) under Germany's Excellence Strategy.}%
\thanks{$^{1}$Learning \& Intelligent Systems Lab, TU Berlin, Germany}%
\thanks{$^{2}$BITS Pilani, India}%
\thanks{$^{3}$RPL, EECS, KTH Royal Institute of Technology, Stockholm, Sweden}%
}
\title{\Huge Multi-modal optimization for manipulation tasks}
\title{\Huge Multimodal k-order Markov Optimization for Asymptotically-Optimal Motion Planning}
\title{\Huge BITKOMO: Combining Sampling and Optimization for Fast Convergence to Optimal Motion Plans}
\title{\Huge BITKOMO: Combining Sampling and Optimization for Fast Convergence in Optimal Motion Planning}
\title{\LARGE \bf
BITKOMO: Combining Sampling and Optimization for Fast Convergence in Optimal Motion Planning
}

\begin{document}

\maketitle
\begin{abstract}
Optimal sampling based motion planning and trajectory optimization are two competing frameworks to generate optimal motion plans. Both frameworks have complementary properties: Sampling based planners are typically slow to converge, but provide optimality guarantees. Trajectory optimizers, however, are typically fast to converge, but do not provide global optimality guarantees in nonconvex problems, e.g. scenarios with obstacles. To achieve the best of both worlds, we introduce a new planner, BITKOMO, which integrates the asymptotically optimal Batch Informed Trees (BIT*) planner with the  K-Order Markov Optimization (KOMO) trajectory optimization framework. Our planner is anytime and maintains the same asymptotic optimality guarantees provided by BIT*, while also exploiting the fast convergence of the KOMO trajectory optimizer. We experimentally evaluate our planner on manipulation scenarios that involve high dimensional configuration spaces, with up to two 7-DoF manipulators, obstacles and narrow passages. BITKOMO performs better than KOMO by succeeding even when KOMO fails, and it outperforms BIT* in terms of convergence to the optimal solution.
\end{abstract}
\section{Introduction}

Generating optimal motions plans is crucial for almost any robotic tasks ranging from typical manipulation tasks such as bin-picking to autonomous navigation of mobile robots. To solve such tasks, the robotics community relies on two powerful motion planning frameworks: Sampling-based planners and trajectory optimization.

Sampling-based planners like RRT*~\cite{karaman2010optimal}, BIT*~\cite{gammell2015batch} or FMT*~\cite{janson2015fast} converge asymptotically to optimal solutions and almost surely provide a solution if one exists~\cite{Lavalle2006}. However, these planners are slow at converging to the optimal trajectory, because improvements to the current best solution only arise when we sample a state nearby~\cite{salzman2016asymptotically}, and often provide non-smooth trajectories that may require post-processing~\cite{geraerts2007creating}.

Trajectory optimization methods like KOMO~\cite{toussaint2014newton}, CHOMP~\cite{ratliff2009chomp}, STOMP~\cite{kalakrishnan2011stomp} and TrajOpt~\cite{schulman2014motion} use optimization methods and can exploit gradient and second order information to converge to a local optimal solution.
These optimization-based methods are typically fast at converging to the local optimum, 
however, due to the non-convexity of the problem, the optimizer might converge to a locally optimal or even an infeasible trajectory. These methods therefore do not have convergence guarantees, i.e. they may not converge to a solution even if one exists, and the feasibility of the solution often depends heavily on the initial trajectory~\cite{merkt2018leveraging, lembono2020memory, ichnowski2020deep}. Hence, these methods usually work well in environments with few obstacles or when provided with good initial guesses~\cite{liu2017planning}, e.g. for post-processing paths produced by sampling-based planners.

\begin{figure}[t]
    \centering
    \includegraphics[width=\linewidth]{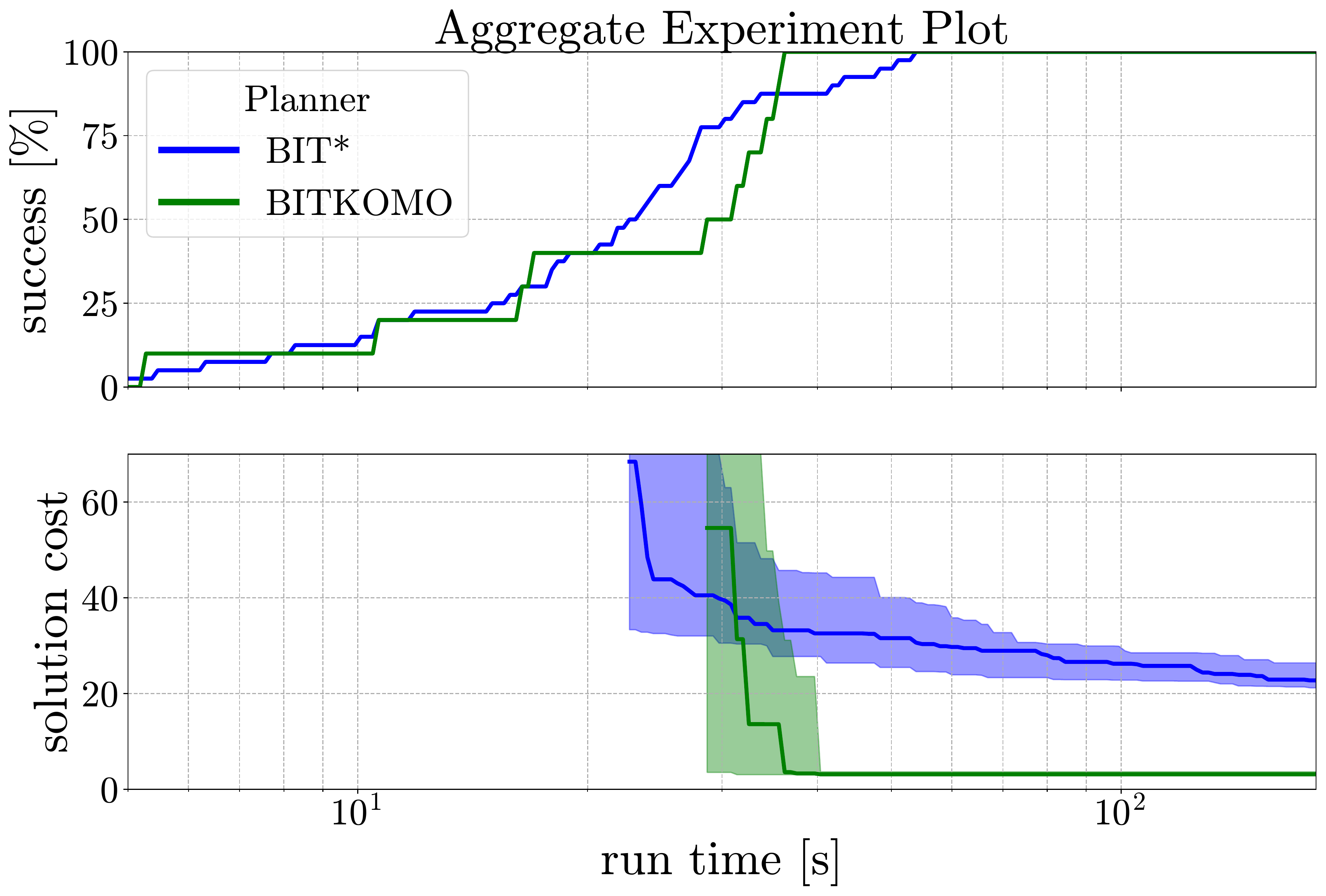}
    \caption{Aggregate plot over all our experiments, showing how BITKOMO converges faster than BIT*, while keeping a similar success rate. KOMO is not included, because it fails to find solutions on more than half of our experiments.}
    \label{fig:pullfigure}
    \vspace*{-0.3cm}
\end{figure}

In order to combine the benefits of both frameworks, we propose to integrate the asymptotically optimal Batch Informed Trees (BIT*)~\cite{gammell2015batch} planner with  K-Order Markov Optimization (KOMO)~\cite{toussaint2014newton, toussaint2015IJCAI}. Combining sampling and optimization helps us play on the strengths of each framework and mitigate their weaknesses. Our novel algorithm, BITKOMO, uses BIT* to iteratively generate non-optimal initial paths that are then optimized by KOMO, which results in quick convergence to local minima. The cost of the optimized path is then used by BIT* to carry a more informed search. With this method, we maintain the asymptotic global optimality guarantees by BIT* while also benefiting from the fast convergence of KOMO (Fig.~\ref{fig:pullfigure}). 

At the core of BITKOMO lies a new relaxed edge collision checking method.
Relaxed edge collision is an intermediate approach between full and lazy~\cite{hauser2015lazy} collision checking, where we allow partially-valid edges to remain, because the trajectory optimizer KOMO can often push invalid paths out of collision~\cite{toussaint2015IJCAI} to converge to feasible solutions. Even though this modification allows for invalid edges, we do not sacrifice any of the asymptotic guarantees provided by BIT*. 
To summarize, we make two major contributions:
\begin{enumerate}
    \item \textit{Relaxed edge collision checking:} A method for BIT* that allows edges partially in collision to be included in the motion tree.
    \item \textit{BITKOMO: A planner integrating BIT* and KOMO.} We integrate sampling and optimization to obtain fast convergence to the global optimum while maintaining the guarantees provided by the sampler.
\end{enumerate}
\section{Related Work}
Combining both asymptotically-optimal planners~\cite{karaman2011sampling, salzman2016asymptotically, gammell2021asymptotically, strub2021ait} and trajectory optimization~\cite{toussaint2014newton, ratliff2009chomp, kalakrishnan2011stomp, schulman2014motion, mukadam2018continuous} into one concise algorithm is an important long-term goal of the robotics community~\cite{hartmann2020planning}. There exist three main approaches. The \emph{two-step approach}, where we run a sampling-based planner like the rapidly-exploring random tree (RRT)~\cite{Lavalle2006}, or the probabilistic roadmap (PRM)~\cite{kavraki1996probabilistic} until a valid solution path is found. In a post-processing step~\cite{kim2003extracting, geraerts2007creating}, we call a trajectory optimizer which optimizes the path to find a (local) optimal solution. This method is the most common approach, but does not hold guarantees on optimality, and does not allow the sampling-based planner to find improved solutions by leveraging information acquired by the optimizer. Our BITKOMO approach improves upon this methodology by tightly integrating sampling and optimization in an iterative fashion. Moreover, we use relaxed collision checking to quickly find partially-valid paths. Optimizers like KOMO~\cite{toussaint2014newton} can often repair those paths to quickly find a solution. 

The second approach is the \emph{optimizer-as-steering} method, where the integration of sampling and optimization is done using the steering function from the planner. The regionally-accelerated BIT* (RABIT*)~\cite{choudhury2016regionally} is an extension of BIT* that uses an optimizer to push infeasible edges out of collision. However, while the obtained solution cost is often better than BIT*, calls to an optimizer are often expensive and slow down the algorithm. It is also possible to integrate such an approach into a roadmap planner~\cite{alwala2020joint}, to interleave edge creation and optimization steps. We differ by postponing optimization to the point where a full partially-valid path has been found, which reduces the number of unnecessary calls to the nonlinear optimizer.

Finally, the \emph{path-proposal} method, where the planner proposes solution paths, which are then sent to the optimizer~\cite{kuntz2020fast, xanthidis2020navigation}. An important aspect of this method is to propose many diverse solution paths, so that the optimizer has a lesser chance to converge to similar solutions. This can be accomplished by leveraging sparse roadmaps~\cite{Orthey2020RAL}, to generate diverse initial trajectories which we can then send to the trajectory optimizer~\cite{dai2018improving, park2015parallel, Orthey2020WAFR}. 
The closest work to ours is~\cite{hartmann2020planning}, where RRT* is integrated with a nonlinear optimizer in an iterative fashion.

Our method is complementary, in that we also use the path-proposal method. However, we differ from previous works by our choice of the underlying components (BIT* and KOMO) and/or the interface. 
Further, the novel collision checking improves the performance 
in problems involving narrow passages, when the sampling-based planner fails to find a path in the first place. 
\section{Problem Definition}
We consider a motion planning problem in a configuration space $\X \subset \R^n$ of the form ($\X_{\text{free}}, x_{\text{start}}, x_{\text{goal}}, c$) where $\X_{\text{free}}\subseteq\X$ is the free configuration space, $x_{\text{start}} \in \X_{\text{free}}$ is the start configuration, $x_{\text{goal}} \in \X_{\text{free}}$ is the goal configuration and $c : \mathcal{P} \to \R$ is the cost functional mapping a trajectory $p \in \mathcal{P}$ in the free configuration space to a real number.
Our goal is to find a trajectory $p : [0,1] \to \X_{\text{free}}$ from $p(0) = x_{\text{start}}$, to $p(1) = x_{\text{goal}}$ that is optimal, i.e. the value of $c(p)$ is the lowest among all possible paths. 

\begin{subequations}
\begin{align}
\min_{p(t)}\quad & c(p(t))\\
        \text{s.t.}\quad & p(0) = x_{\text{start}}, \hspace{5px}  p(1) = x_{\text{goal}}\\
        & \forall t \in [0,1], \hspace{5px} p(t) \in \X_{\text{free}} ~.
\end{align}
\end{subequations}
In this paper, we focus on path-length optimization, i.e.  $c(p) = \int_0^1{||\dot{p}(t)||~dt}$. Typical examples of alternative cost functionals are  the sum of the velocities squared  $\int_0^1{||\dot{p}(t)||^2~dt}$, and smoothness $\int_0^1{||\ddot{p}(t)||^2~dt}$.
\section{BITKOMO}

BITKOMO integrates two state-of-the-art motion planners: BIT* and KOMO. BIT* is an anytime, asymptotically-optimal planner that samples collision-free configurations in batches and generates paths using the A* graph search algorithm. KOMO is a non-linear trajectory optimizer that locally optimizes an initial trajectory (possibly in collision) to minimize a cost and fulfil collision avoidance and goal constraints. 
Our planner maintains the asymptotic optimality guarantees of BIT* while converging faster to the global minimum by leveraging trajectory optimization. Since KOMO uses the Augmented Lagrangian algorithm for constrained optimization, it can sometimes push partially invalid paths out of collision. To exploit this feature, we introduce relaxed edge checking, which allows BIT* to produce partially infeasible paths for subsequent optimization with KOMO when necessary.

\subsection{The BITKOMO Algorithm: An overview}
\begin{figure}
    \centering
    \includegraphics[width=0.9\linewidth]{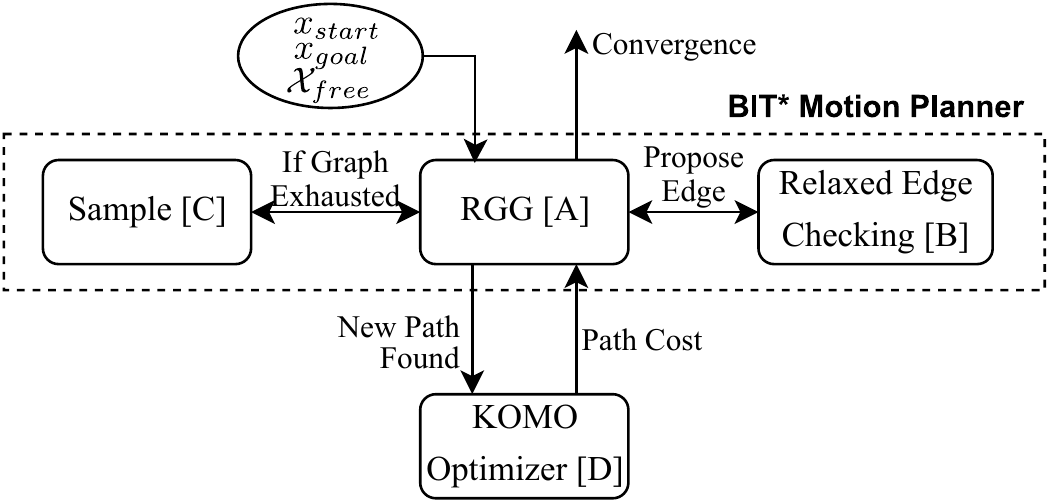}
    \caption{Overview of BITKOMO, which combines the BIT* architecture with our custom relaxed edge checking for path planning, and the KOMO optimizer for path optimization.}
    \label{fig:overview}
\end{figure}

Our planner (Fig. \ref{fig:overview}) requires a valid start state ($x_{\text{start}}$), a goal state ($x_{\text{goal}}$) and the full information about the environment ($\X_{\text{free}}$). We also need to provide the Planner Termination condition ($PTC$) and the edge relaxation number ($\delta$). To begin, the BIT* planner samples a batch of configurations $x \in \mathcal{X}_{\text{free}}$ and builds an \textit{edge-implicit} Random Geometric Graph (RGG)~\cite{penrose2003random}~[Block A in Fig. \ref{fig:overview}]. The best edge that can possibly improve the cost to goal (as in A*) is then chosen and passed to the \textit{Relaxed Edge Checker}~[B] to carry out the collision checking. The Relaxed Edge Checker performs a validity check and returns the collision penalty ($\collisionPenalty$), an integer that provides a proxy measure for the fraction of the edge that is in collision. The planner uses this integer to decide regarding the addition of the edge to the tree. If an improved path to the goal is found, it is passed to the KOMO optimizer~[D] which locally optimizes the path and, if valid, returns the new cost to the BIT* planner. BIT* uses this path cost to prune the unnecessary vertices and edges and carry out a more focused search. When no new edges can be expanded, the Sample function~[C] is called, which adds another batch of samples to the RGG.

\begin{algorithm}
\caption{BITKOMO}
\hspace*{\algorithmicindent} \textbf{Input} $\X_{\text{free}},x_{\text{start}},x_{\text{goal}},PTC,\delta$
\begin{algorithmic}[1]
    \State $\mathcal{V} \gets \{x_{\text{start}}\}$; $\mathcal{E} \gets \emptyset$; $\tree \gets (\mathcal{V},\mathcal{E})$; \tikzmark{top1}
    \State $\vertexQueue \gets V$; $\edgeQueue \gets \emptyset$;
    \State $\unconnectedSet \gets x_\text{goal}$; 
    \State $\bestCost \gets \infty$; $\bestCostBITKOMO \gets \infty$;
    \State $\maxCost \gets \Call{GetDiagonalLength}{\X} \times 3$; \label{Algo:BITKOMO:maxCost} \tikzmark{bottom1}
    \While{$\neg PTC$}
        \If{$\edgeQueue \equiv \emptyset$ and $\vertexQueue \equiv \emptyset$}
        \label{Algo:BITKOMO:BathAdditionCondition}
        \tikzmark{top2}
            \State $\unconnectedSet \adds \Call{Prune\&Sample}{\tree, \unconnectedSet, m, \bestCost}$;
            \State $\vertexQueue \gets V$;
        \EndIf \tikzmark{bottom2}
        \While{$\Call{BestValue}{\vertexQueue} \le \Call{BestValue}{\edgeQueue}$} \tikzmark{top3}
            \State $\Call{ExpandNextVertex}{\vertexQueue,\edgeQueue,\bestCost}$; \label{Algo:BITKOMO:ExpandNextVertex}
        \EndWhile
        \State $E = \{v_\text{min},x_\text{min}\} \gets \Call{PopBestInQueue}{\edgeQueue}$; \tikzmark{bottom3}
        \If{$\Call{EdgeAdditionHelps}{E,\bestCost}$} \tikzmark{top4} \label{Algo:BITKOMO:EdgeAdditionHelps}
            \color{blue}
            \State $\collisionPenalty \gets \Call{CheckEdgeRelaxed}{E}$; \label{Algo:BITKOMO:RelaxedCollisionChecking}
            \If{$\collisionPenalty \le \delta$} \Comment{If true, Edge is used\tikzmark{right}} \label{Algo:BITKOMO:collisionThreshold}
                \State $c_\text{edge} \gets \hat{c}(E) + \collisionPenalty \times \maxCost$; \label{Algo:BITKOMO:AddingCollisionCost}
            \color{black}
                \If{$\Call{EdgeImprovesCost}{E,\bestCost, c_{\text{edge}}}$} \label{Algo:BITKOMO:EdgeImprovesCost}
                    \State $\Call{AddEdgeToTree}{E}$; \tikzmark{bottom4} \label{Algo:BITKOMO:AddEdgeToTree}
                    \State $\bestCost \gets \Call{GetCostToGo}{v_{\text{goal}}}$; \tikzmark{top5} \label{Algo:BITKOMO:updateBestCost}
                \color{blue}
                    \If{$\bestCost < \maxCost$} \label{Algo:BITKOMO:isPathFeasible} \Comment{If path is feasible}
                        \State $\bestCostBITKOMO \gets \bestCost$; \label{Algo:BITKOMO:setBestCost1}
                    \EndIf
                    \color{black}
                    \State $path \gets \Call{GetBestPath}{v_{\text{goal}}}$;
                    \color{orange}
                    \State $optiPath \gets \Call{Optimize}{path}$; \label{Algo:BITKOMO:optiPath}
                    \If{$\Call{isValid}{optiPath}$}
                        \State $\bestCost \gets \Call{GetCost}{optiPath}$; \label{Algo:BITKOMO:updateBestCost2}
                        \State $\bestCostBITKOMO \gets \bestCost$; \label{Algo:BITKOMO:updateBestCostBITKOMO}
                    \EndIf \tikzmark{bottom5}
                    \color{black}
                \EndIf
            \EndIf
        \EndIf
    \EndWhile
    \State \Return Solution;
\end{algorithmic}
\AddNote{top1}{bottom1}{right}{\textit{A}}
\AddNote{top2}{bottom2}{right}{\textit{B}}
\AddNote{top3}{bottom3}{right}{\textit{C}}
\AddNote{top4}{bottom4}{right}{\textit{D}}
\AddNote{top5}{bottom5}{right}{\textit{E}}
\label{Algo:BITKOMO}
\end{algorithm}

Algorithm \ref{Algo:BITKOMO} describes in more detail the different parts of the planner. The highlighted lines are our addition to the BIT* planner. Blue --- the \textit{Relaxed edge collision checking}, Orange --- the interface between BIT* and KOMO. 

\subsubsection{Initialize (A)} \label{AlgoDef:Initialize} Vertex set ($\mathcal{V}$), Edge set ($\mathcal{E}$), Tree ($\tree$), Vertex queue ($\vertexQueue$), Edge queue ($\edgeQueue$), Set of unconnected vertices ($\unconnectedSet$). Also initialize three important cost parameters:  1) $\bestCost$ --- cost used by the BIT* tree, it includes infeasible paths; 2) $\bestCostBITKOMO$ --- the cost of the best feasible path; 3) $\maxCost$ --- a penalty cost higher than any feasible path BIT* could converge to.

\subsubsection{Batch Addition (B)}\label{AlgoDef:BatchAddition} When we run out of the batch samples (line~\ref{Algo:BITKOMO:BathAdditionCondition}), we prune our graph using the ellipsoid method and add a new batch of samples~\cite{gammell2020batch}.

\subsubsection{Edge Selection (C)} \label{AlgoDef:EdgeSelection} (The A* search) If expanding a vertex can help improving the cost of our solution, it is expanded, i.e., relevant vertices and edges are added to their respective queues (line~\ref{Algo:BITKOMO:ExpandNextVertex})

\subsubsection{Edge processing (D)} \label{AlgoDef:EdgeProcessing} Decides whether to add new edge to the tree. If the new edge can improve the overall cost to goal (line~\ref{Algo:BITKOMO:EdgeAdditionHelps}), and the edge is collision free / partially in collision (line~\ref{Algo:BITKOMO:RelaxedCollisionChecking},~\ref{Algo:BITKOMO:collisionThreshold}) such that it still is a good edge to add (line~\ref{Algo:BITKOMO:EdgeImprovesCost}), it is added to the tree (line~\ref{Algo:BITKOMO:AddEdgeToTree}). $\Call{AddEdgeToTree}{.}$ rewires the tree if necessary.

\subsubsection{KOMO Optimization (E)} \label{AlgoDef:KOMOOptimization} If the addition of the new edge provides us with a better path to goal, this solution path is optimized using KOMO (line~\ref{Algo:BITKOMO:optiPath}). The resulting path~($optiPath$) is then checked for validity and the costs ($\bestCost$ and $\bestCostBITKOMO$) are updated accordingly. For completeness, we also check if the initial guess is valid by checking if the path cost is less than $\maxCost$ (line~\ref{Algo:BITKOMO:isPathFeasible}) and update $\bestCostBITKOMO$ if valid. The working of KOMO is explained in \ref{KOMO}.

\begin{figure}
    \centering
    \includegraphics{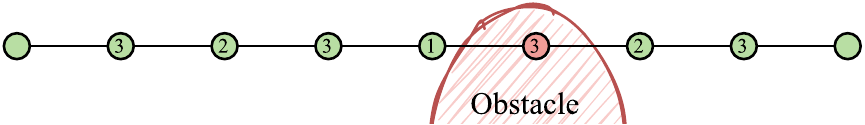}
    \caption{\textit{Levels in Edge Checking} The edge is first subdivided into $n_d$ points and these points are then checked for collision level by level. The number on the nodes represent the level of the node. This example has 3 levels.}
    \label{fig:levels}
\end{figure}

\subsection{Relaxed Edge Checking}
High dimensional spaces containing narrow passages are challenging for sampling based planners. This is because it is difficult to sample collision free edges through narrow passages.
Since KOMO can push paths out of obstacles, we could allow paths partially in collision into the BIT* tree. However, these edges need to be added with sufficient collision penalty to ensure that BIT* does not mistake a path in collision to be of a lower cost than the true minima. 
We also want our collision checker to quickly guess the extent of collision so as to be quick in finding a solution for BIT*.
We solve this problem by introducing Relaxed Edge Checking which returns a number instead of a Boolean which is used to assign a collision penalty (line~\ref{Algo:BITKOMO:AddingCollisionCost}).
It returns 0-if edge is collision free, 1-if it fails at the last level, 2-if it fails on the second to last level, and so on.
Adding the collision penalty this way also helps our planner to prefer collision free initial paths for optimization as the likelihood of finding a feasible trajectory from a collision-free path is higher.

Suppose for a given resolution, we need to check $n_d$ equally spaced points to confirm the edge to be collision-free. 
The Relaxed Edge Checker conducts a level wise collision checking (see Fig.~\ref{fig:levels}) whereby the resolution of checking is increased until the required resolution is reached or a collision is detected. We first check the mid point (level 1), then the quarter points (level 2) and so on by slowly doubling the resolution of checking. If a point fails in the validity check, an integer, collision penalty~($\collisionPenalty = \level - \level_c + 1$) is returned. Where $\level = \ceil{\log_2n_d}$ is the total number of levels, and $\level_c$ is the level of the failed point. This number provides a proxy measure for the fraction of the edge that is in collision.

\subsection{KOMO} \label{KOMO}
K-Order Markov Optimization (KOMO) is a trajectory optimization framework that represents a path with a discrete sequence of waypoints  $\langle x_0 \ldots x_T \rangle$. Cost and constraints are evaluated on, up to $k+1$ consecutive waypoints (Markov assumption)
\begin{subequations}
\label{eq:KOMO}
\begin{align}
    \min_{x_{0:T}}\quad & \sum_{t=0}^T f_t(x_{t-k:t})^\top f_t(x_{t-k:t})\\
    \text{s.t.}\quad &\forall t : g_t(x_{t-k:t}) \le 0, \hspace{10pt} h_t(x_{t-k:t}) = 0 ~,
\end{align}
\end{subequations}
where $x_{t-k:t}$ is a $k+1$ tuple of consecutive states. In our setting, where the goal is to minimize the path length, $k=1$, and we use, as cost, the sum of squared distances $\sum || x_t - x_{t-1} ||^2$, which corresponds to  $f_t(x_{t-1},x_{t}) = x_{t}-x_{t-1}$.  

Inequality constraints correspond to collision avoidance and joint limits and equality constraints model the terminal goal condition $x_T=x_{\text{goal}}$. 
The optimization problem \eqref{eq:KOMO} is solved with the Augmented Lagrangian algorithm for constrained optimization. The Markov structure, together with second order information, enables very efficient solving, with complexity linear on the number of waypoints and polynomial on the dimension of the configuration space \cite{toussaint2014newton}.

\subsection{Convergence and Optimality Guarantees} 

BITKOMO maintains the convergence and optimality guarantees of BIT*~\cite{gammell2015batch}. The additional trajectory optimization can only improve the solution proposed by BIT* (lines 26-30 in Alg. \ref{Algo:BITKOMO}).
The \textit{relaxed edge checking} assigns cost $c>c_{\text{max}}$ (line 18 in Alg. \ref{Algo:BITKOMO}) to any edge in collision (recall that $c_{\text{max}}$ is an upper bound on the optimal solution cost, that can be chosen arbitrarily large). 
Even if the subsequent optimization fails, the edge cost does not prevent BIT* and hence BITKOMO from finding a solution with cost $c<c_{\text{max}}$. 

\section{Evaluation}
\begin{figure}[t]
    \begin{subfigure}{.3\linewidth}
        \centering
        \includegraphics[width=\linewidth]{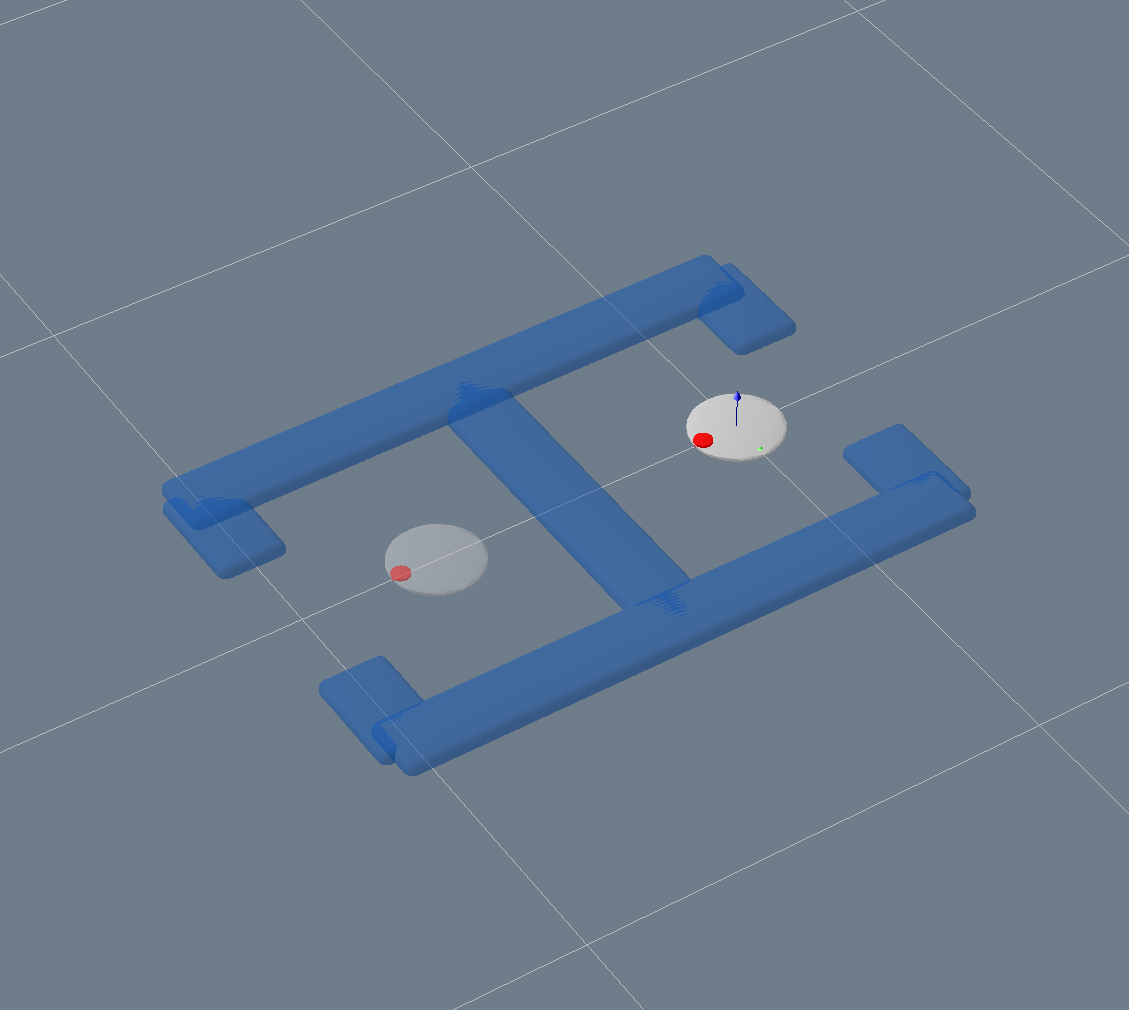}
        \caption{Disc Robot}
        \label{fig:Scene_DiscRooms}
    \end{subfigure}
    \begin{subfigure}{.3\linewidth}
        \centering
        \includegraphics[width=\linewidth]{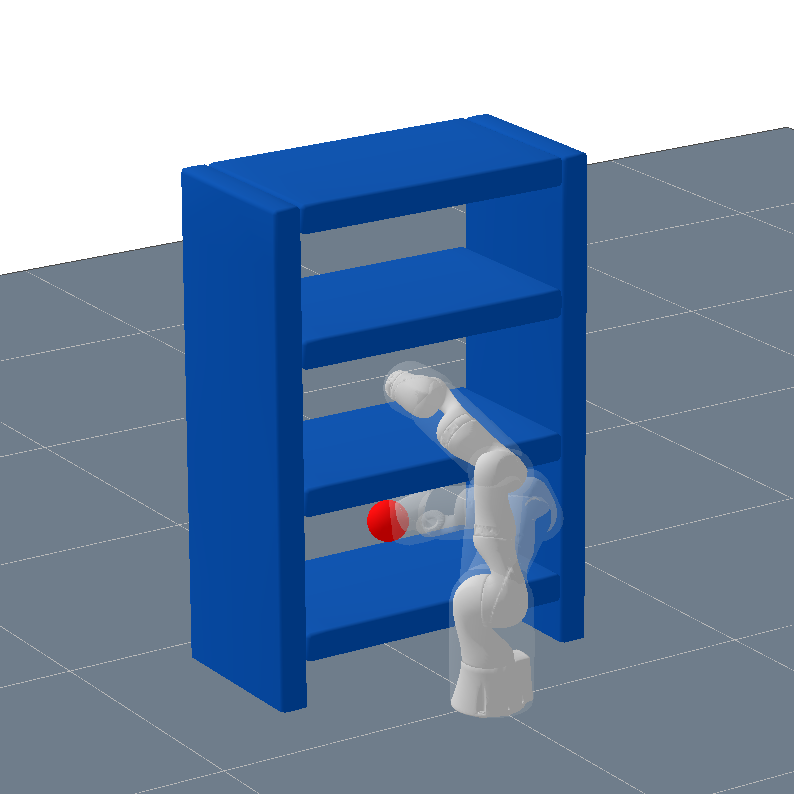}
        \caption{Kuka from shelf}
        \label{fig:Scene_KukaShelf}
    \end{subfigure}
    \begin{subfigure}{.3\linewidth}
        \centering
        \includegraphics[width=\linewidth]{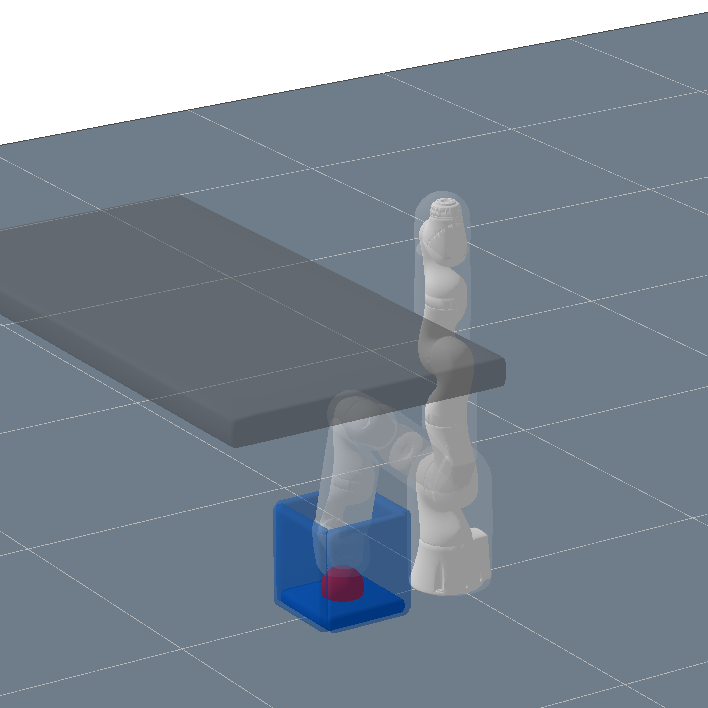}  
        \caption{Kuka into box}
        \label{fig:Scene_KukaBox}
    \end{subfigure}
    
    \begin{subfigure}{.3\linewidth}
        \centering
        \includegraphics[width=\linewidth]{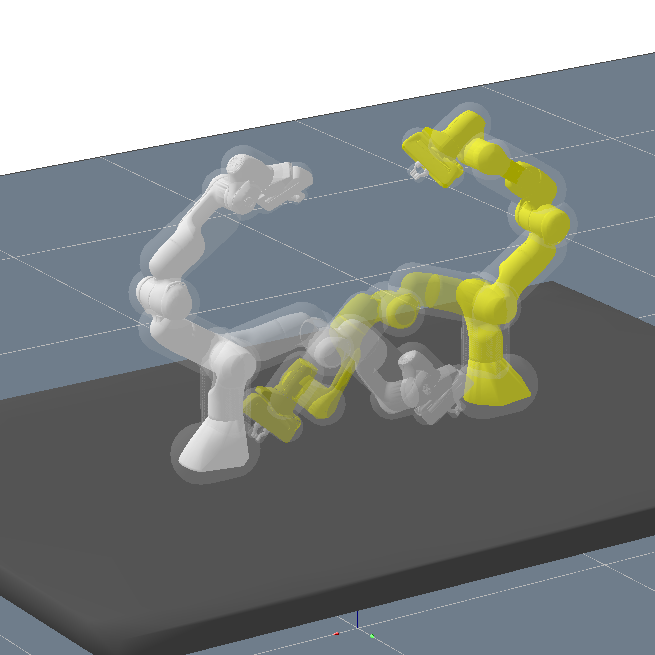}
        \caption{Fixed Pandas}
        \label{fig:Scene_Pandas}
    \end{subfigure}
    \begin{subfigure}{.3\linewidth}
        \centering
        \includegraphics[width=\linewidth]{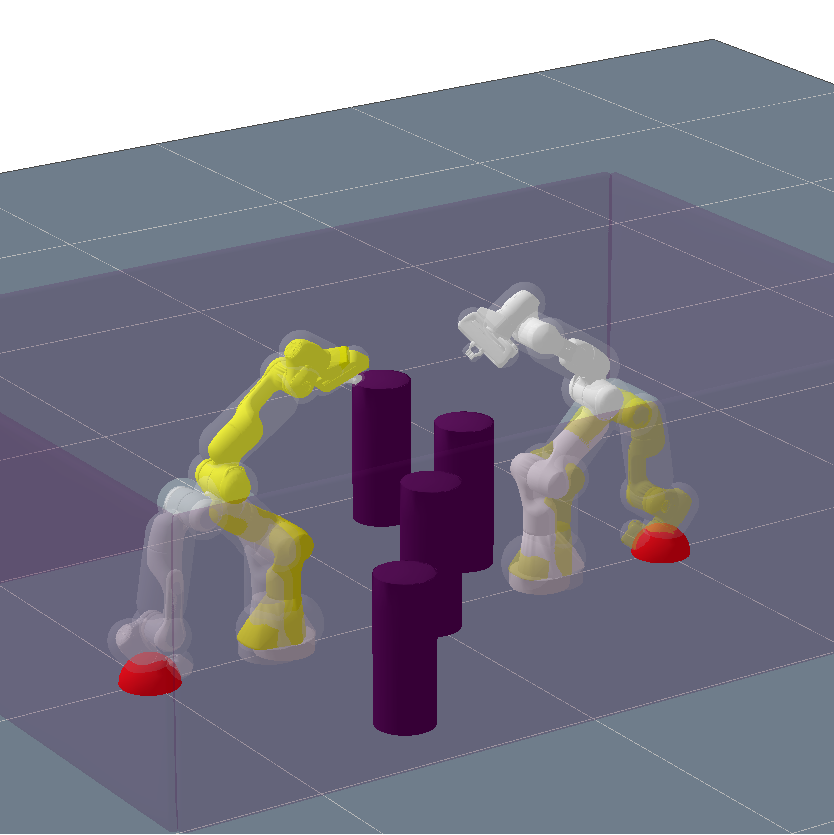}
        \caption{Two Mobile Pandas}
        \label{fig:Scene_twoMobilePandas}
    \end{subfigure}
    \begin{subfigure}{.3\linewidth}
        \centering
        \includegraphics[width=\linewidth]{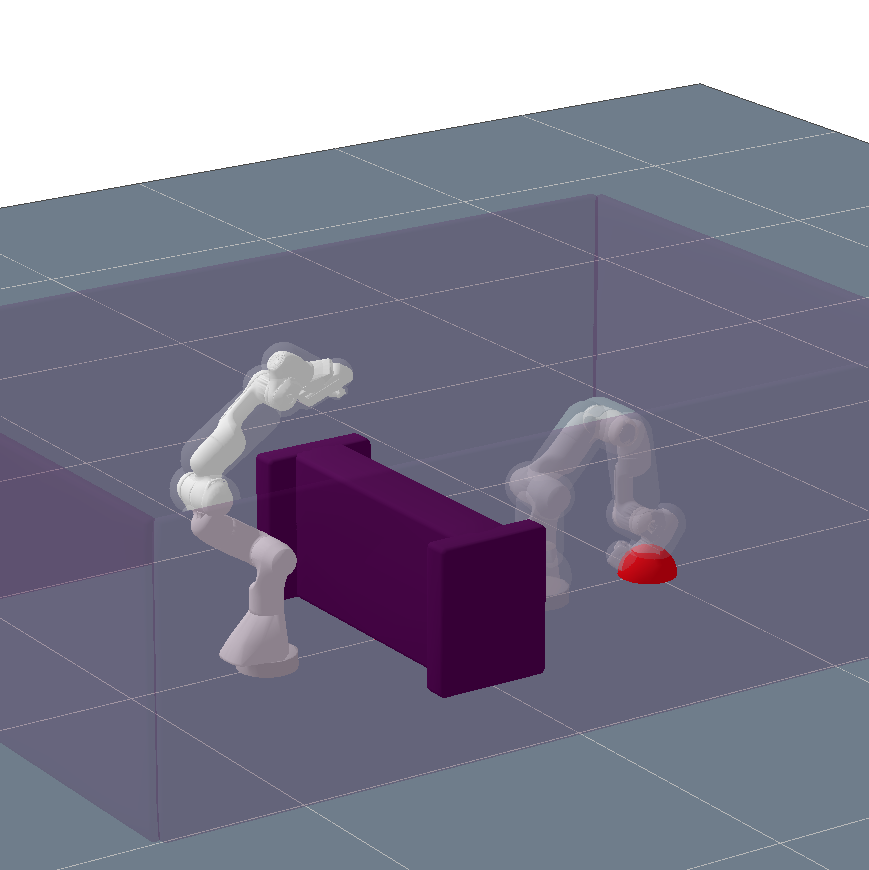}  
        \caption{One Mobile Panda}
        \label{fig:Scene_oneMobilePanda}
    \end{subfigure}
    \caption{Scenarios used in our experimental evaluation. See the supplementary video for the solution trajectories.}
    \label{fig:Scenarios}
\end{figure}
\begin{figure*}[t]
    \def\figWidth{0.32\linewidth}
    \begin{subfigure}{\figWidth}
        \centering
        \includegraphics[width=\linewidth]{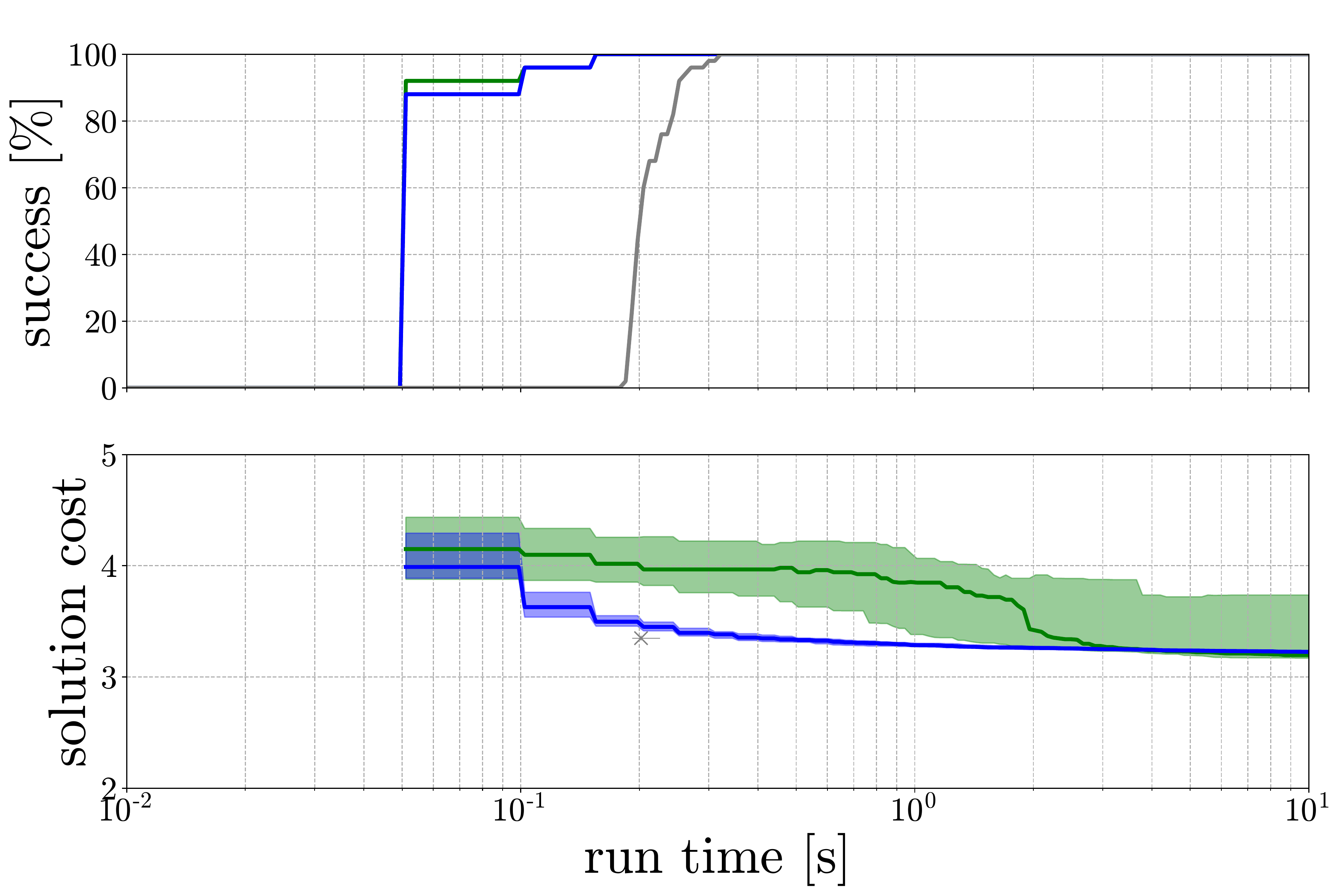}  
        \caption{Disc Robot in Rooms}
        \label{fig:Disc_rooms}
    \end{subfigure}
    \begin{subfigure}{\figWidth}
        \centering
        \includegraphics[width=\linewidth]{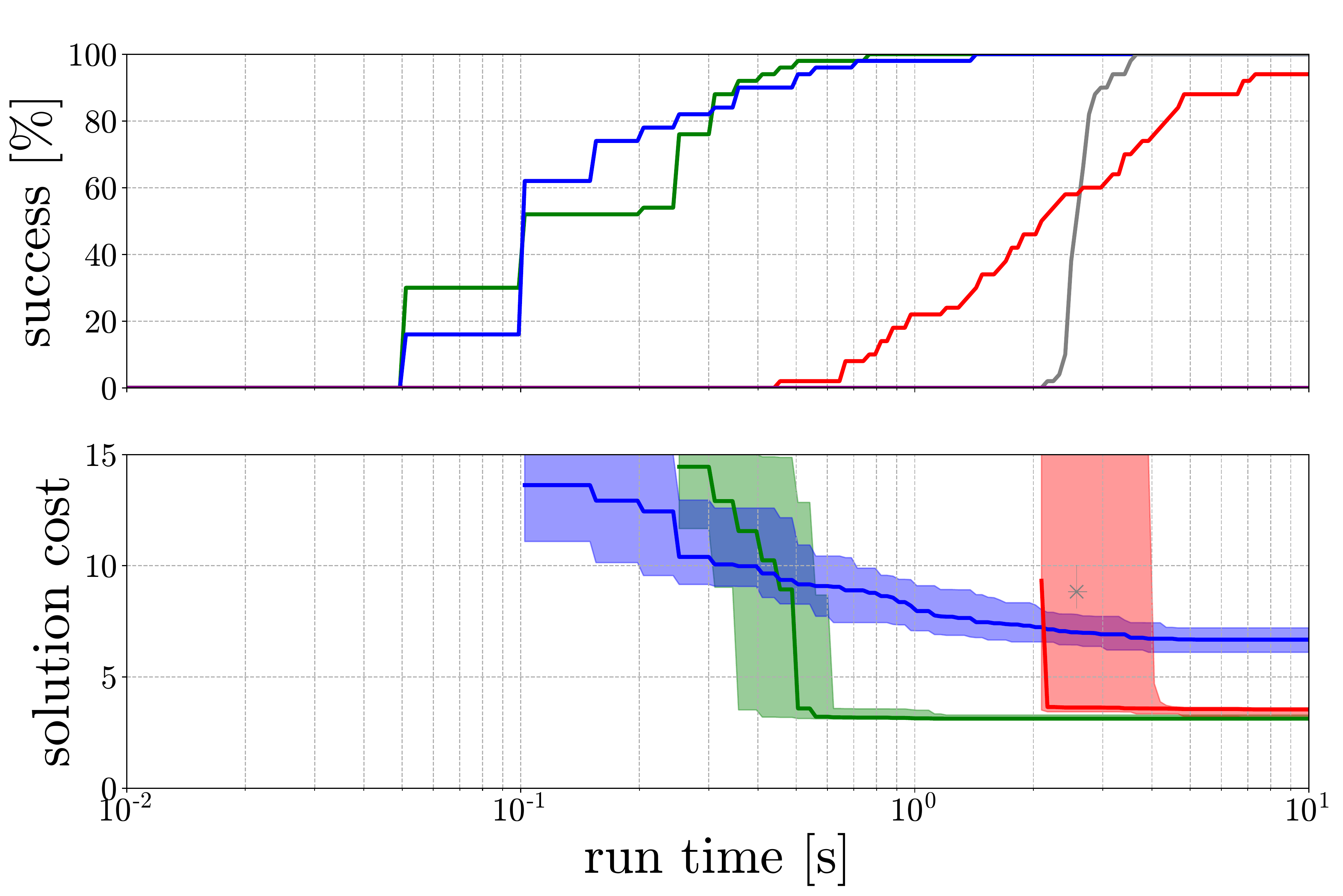}  
        \caption{Kuka from shelf}
        \label{fig:kuka_shelf}
    \end{subfigure}
    \begin{subfigure}{\figWidth}
        \centering
        \includegraphics[width=\linewidth]{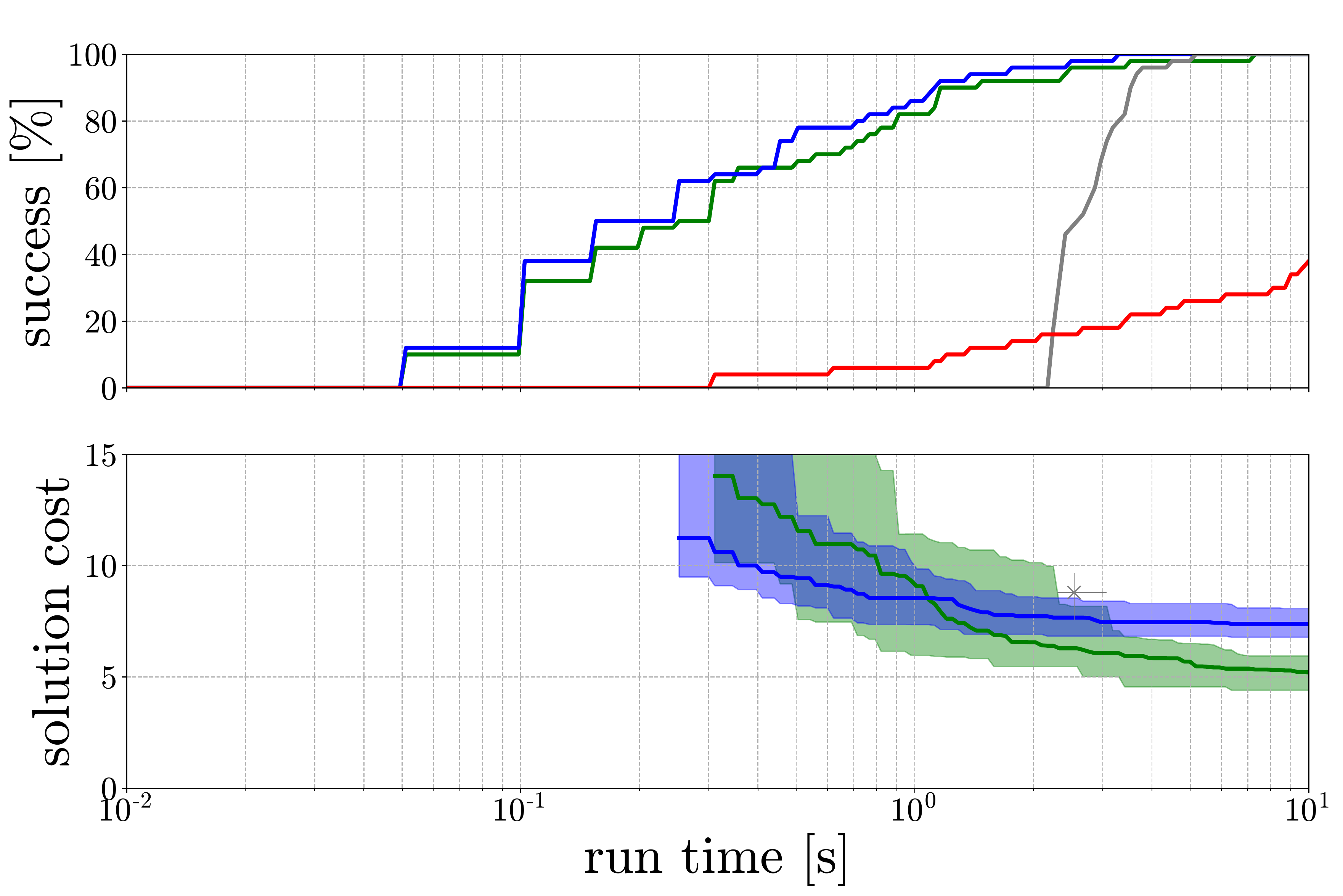} 
        \caption{Kuka into the box}
        \label{fig:kuka_box}
    \end{subfigure}
    
    \begin{subfigure}{\figWidth}
        \centering
        \includegraphics[width=\linewidth]{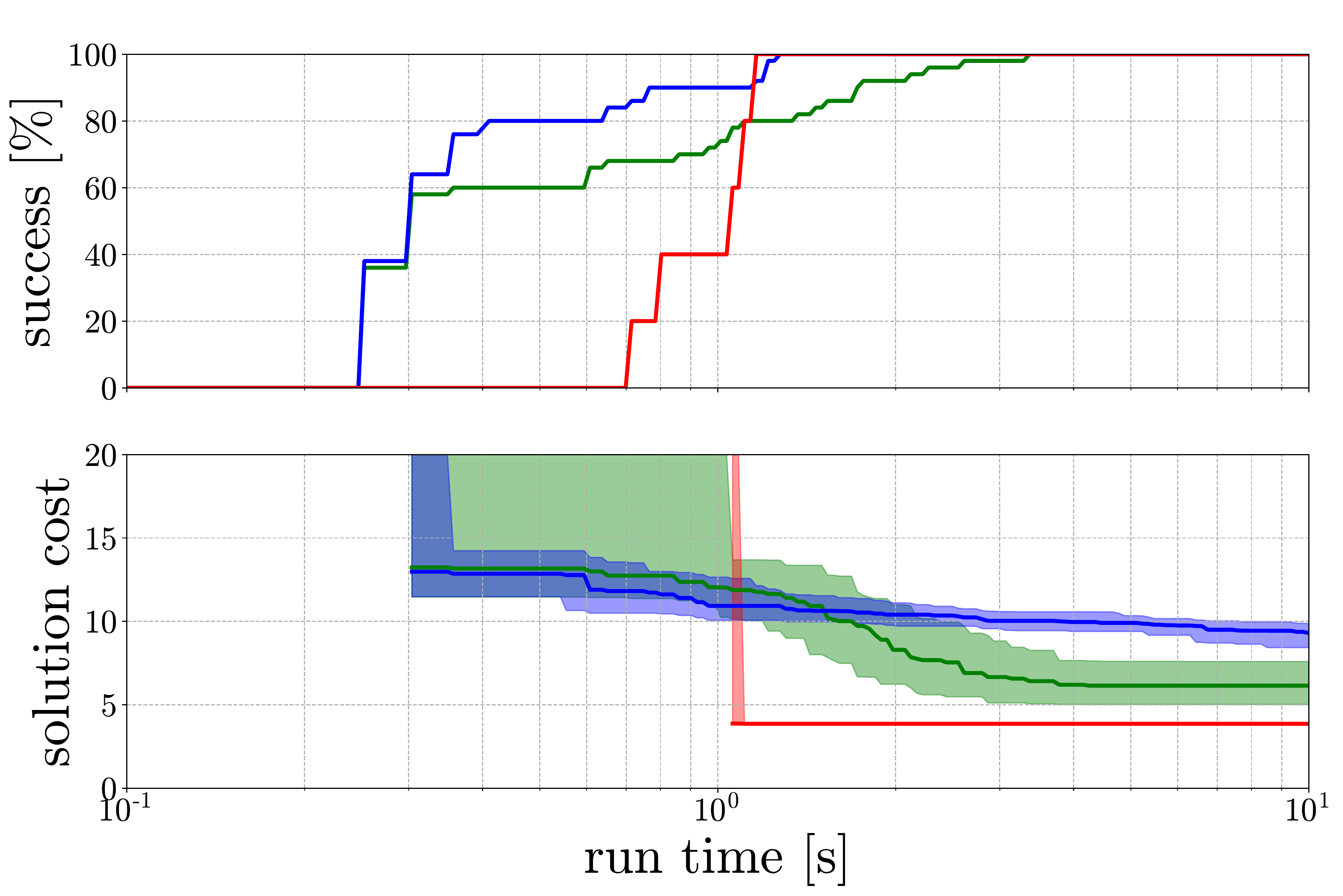}  
        \caption{Fixed Pandas}
        \label{fig:Two_Pandas}
    \end{subfigure}
    \begin{subfigure}{\figWidth}
        \centering
        \includegraphics[width=\linewidth]{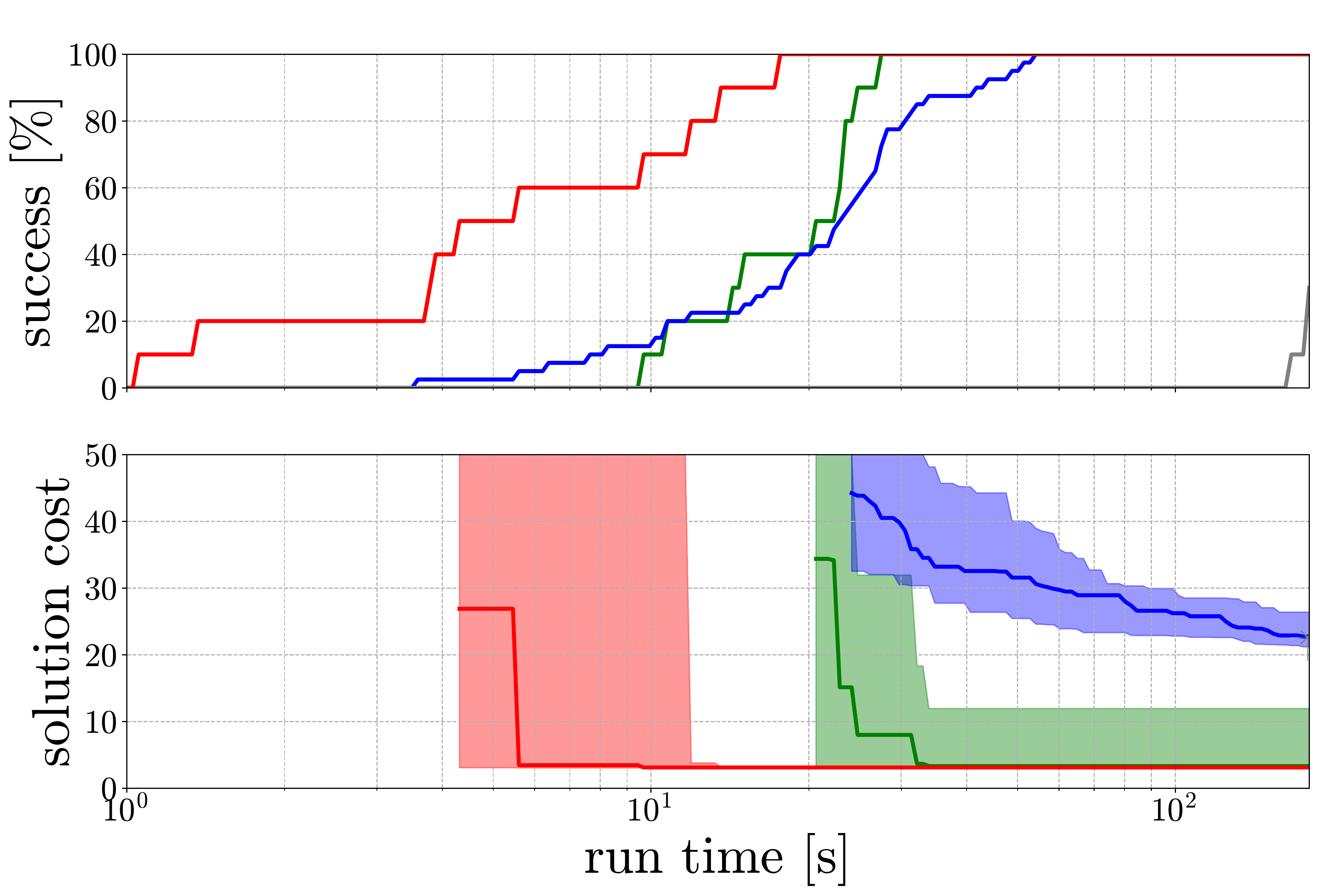}
        \caption{Two Mobile Pandas}
        \label{fig:TwoMobileManipulators}
    \end{subfigure}
    \begin{subfigure}{\figWidth}
        \centering
        \includegraphics[width=\linewidth]{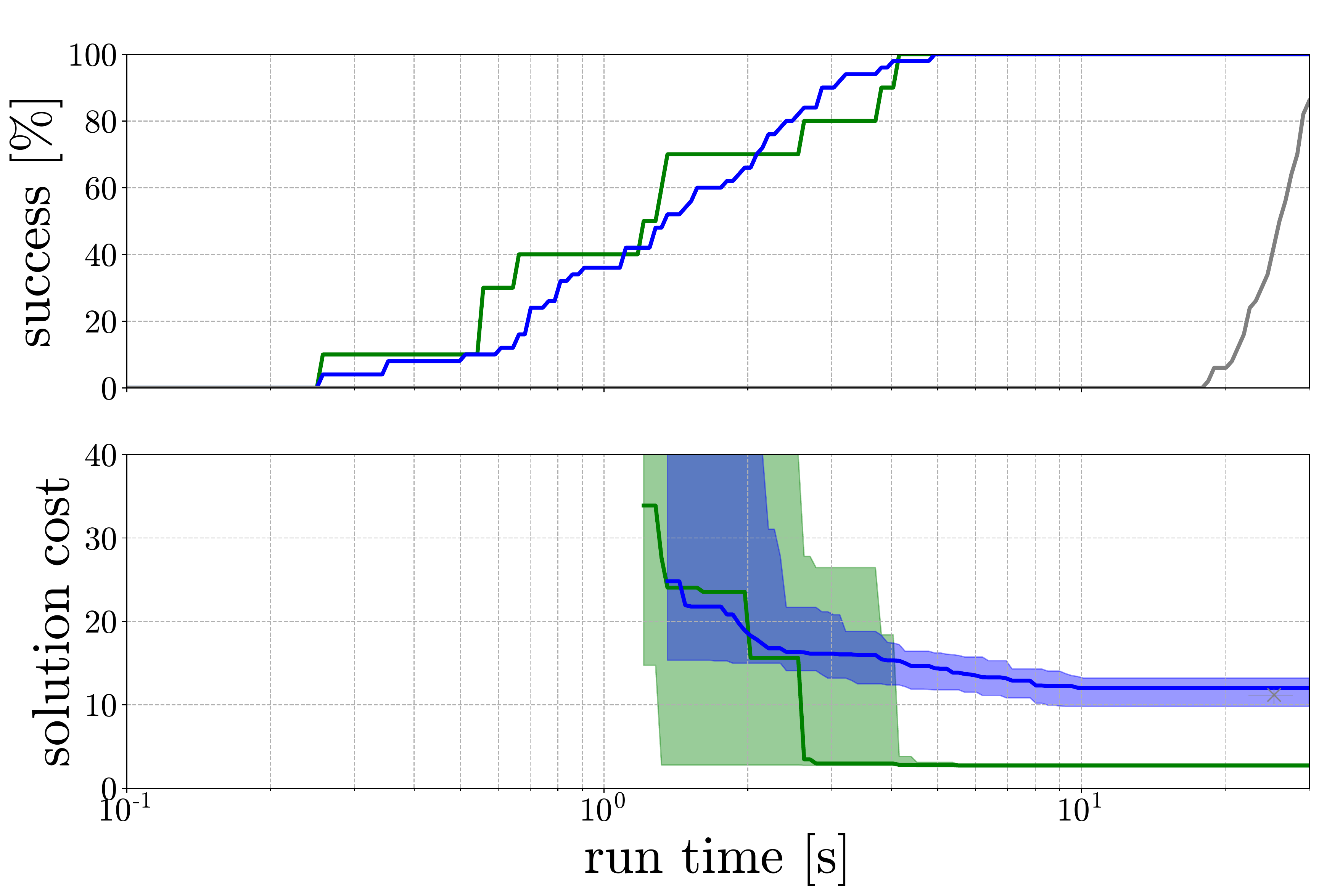}
        \caption{One Mobile Panda}
        \label{fig:OneMobilePanda}
    \end{subfigure}
    
    \newcommand\cbox[1]{\raisebox{0.1cm}{\colorbox{#1}{}}}
    
    \caption{Results: Success rates and best cost plots for \cbox{bitkomogreen} BITKOMO, \cbox{blue!50} BIT*, \cbox{gray!50} FMT* and \cbox{red!50} KOMO on the 6 different example environments.
\vspace*{-0.5cm}    
    }
    \label{fig:benchmarks}
\end{figure*}

\subsection{Scenarios}

We evaluate our algorithm on 6 different robotic scenarios\footnote{ \href{https://github.com/JayKamat99/mt-multimodal_optimization/tree/IROS_2022}{https://github.com/JayKamat99/mt-multimodal\_optimization/tree/IROS\_2022}}. In all scenarios, the robot moves from the initial configuration (solid color) to the goal configuration (translucent color)  (Fig.~\ref{fig:Scenarios}). The trajectories computed by BITKOMO and the baseline algorithms are shown in the supplementary video\footnote{\href{https://www.youtube.com/watch?v=HveYWl4wMAI}{https://www.youtube.com/watch?v=HveYWl4wMAI}}. 
We emphasize the challenges of each problem with the keywords: \textit{narrow passage}, \textit{not informative heuristic} and \textit{high-dimensional}. 

\begin{enumerate}
    \item \textit{Disc Robot in Rooms:} A Disc Robot needs to move from the center of one room to another (Fig. \ref{fig:Scene_DiscRooms}). The difficulty is that, to go to the other room, the robot first needs to come out of the starting room and then move to the target room. Challenges: narrow passages and not informative heuristic.
    \item \textit{Kuka to reach onto the shelf:} The Kuka robot needs to reach the red object at level 1 from it's current position where the end-effector is at level 2 (Fig.\ref{fig:Scene_KukaShelf}).
    Challenges: High-dimensional.
    \item \textit{Kuka to reach into a box:} The Kuka Robot needs to reach to the object inside the box that is located under a table while avoiding collision with the table or the box (Fig. \ref{fig:Scene_KukaBox}). Challenges: Narrow Passage, High Dimensional.
    \item \textit{Two Fixed Pandas:} The robotic manipulators (Pandas) need to get to the base of the opposite robot while avoiding hitting each other (Fig. \ref{fig:Scene_Pandas}). Challenges: High Dimensional.
    \item \textit{Two Mobile Pandas in cluttered environment:} Two mobile panda robots need to move to the other side of the room while avoiding obstacles and also avoiding each other (Fig. \ref{fig:Scene_twoMobilePandas}). Challenges: High Dimensional, Narrow Passages.
    \item \textit{One Mobile Panda to avoid large obstacle:} The mobile panda needs to move to catch an object on the other side of the scene, with a large obstacle blocking it's way (Fig. \ref{fig:Scene_oneMobilePanda}). Challenges: High Dimensional, Narrow Passages.
\end{enumerate}

\subsection{Baselines}
We compare our BITKOMO planner with BIT*~\cite{gammell2015batch}, KOMO~\cite{toussaint2014newton} and FMT*~\cite{janson2015fast}.
Path length minimization is used as the optimization objective for all the experiments. We used the Open Motion Planning Library (OMPL)~\cite{sucan2012open} framework for the implementations of the sampling based planners and for carrying out the benchmarks. 

For the KOMO planner we use the sum of squares of the distances between waypoints as optimization objective, and an initialization with random noise around  $p(t_i) = x_{\text{start}} ~ \forall t_i$.
The trajectory is represented with a constant number of waypoints (20 points). 
Random initialization and optimization are executed iteratively until timeout, updating the path cost when a better path is found. 

\subsection{Metrics}
We evaluate the planners on 2 different metrics:
\begin{enumerate}
    \item \textit{Success rate (\%):} The \% of runs that have found a feasible solution at time $t$. This metric gives information about how fast the planner finds the first feasible path.
    \item \textit{Cost:} The average best cost of the planner at time $t$. This metric gives us an understanding about how the best cost solution of a planner evolves over time and the practical convergence speed before the timeout.
\end{enumerate}

\subsection{Experimental Results}

For getting unbiased results, all experiments were conducted on the same machine\footnote{Intel(R) Core(TM) i5-6200U CPU @ 2.30GHz having 16GB RAM}. Every planner was executed 50 times on all the six example scenarios. The maximum execution time however was different for different scenarios depending on the difficulty. The edge relaxation number $\delta$ was set to 1 for all examples.
The results of the benchmarks are shown in Fig. \ref{fig:benchmarks}. 

\subsubsection{Success rate} The success rate of BIT* and BIKOMO were higher compared with other planners in all examples except the \textit{Two Mobile Pandas} example. This scenario has narrow passages which makes it hard for sampling based planners, however, the optimal solution is very similar to a straight line path in the configurations space, making it very easy for KOMO to find a solution here. The relaxed edge checking helps BITKOMO in having a slightly better success rate than BIT* here.
The anomaly in BIKOMO success rate in Fig.~\ref{fig:Two_Pandas} is because of a failed optimization. This failure is because of the thin obstacles in the C-Space arising due to collision between robots. This, however, is not a large time difference. Choosing a smaller edge relaxation number, $\delta$, will fix it.
Overall, the success rates of BIT* and BITKOMO were found to be very similar because they generate initial paths using the same base algorithm.

\subsubsection{Cost} BITKOMO decreases the cost significantly faster than BIT*, with better convergence before the timeout. This is because the combination of sampling and optimization converges to the local minima  quickly and consistently. We however see an abnormality in Fig.\ref{fig:Disc_rooms}. This is because --- 1) KOMO is not much faster than sampling for low dimensions, and 2) The waypoints maintain a certain minimum distance from the obstacles to avoid edge collisions.

Overall, we conclude that our planner is mostly as good as BIT* in finding the first feasible solution and slightly faster in high dimensional narrow passage problems, but much faster at converging to the global optimal solution.
\section{Discussion and Conclusion}
Our planner, BITKOMO, combines BIT* and KOMO to achieve fast convergence to the optimal solution while being anytime and asymptotically converging to the global minimum. Our experiments indicate that BITKOMO converges to the global optima, faster than BIT*.
It also provides convergence guarantees which KOMO does not. Using Relaxed Edge Checking, our planner exploits the ability of KOMO to move trajectories away from the obstacles that are in collision by allowing partially infeasible paths as initial guesses to the optimizer. This helps BITKOMO find motions through narrow passages faster than BIT*. 



Even though we observe faster convergence than BIT* to optimal paths, our planner does not have a better success rate.
A dedicated planner could be developed to generate improved initial guesses to the optimizer, resulting in an increased success rate.
The optimization and sampling modules could also easily be parallelized, providing a higher speed-up.
Calling the KOMO optimizer ahead of the BIT* planner could increase the speed further. 

Our experiments clearly demonstrate that BITKOMO can robustly achieve fast convergence to optimal motion plans. This is an important step towards making optimal motion planners converge as quickly as trajectory optimizers --- all while keeping asymptotic optimality guarantees.

\bibliographystyle{IEEEtran}
{
\balance
\small
\bibliography{bib/general}

\begin{thebibliography}{10}
\providecommand{\url}[1]{#1}
\csname url@samestyle\endcsname
\providecommand{\newblock}{\relax}
\providecommand{\bibinfo}[2]{#2}
\providecommand{\BIBentrySTDinterwordspacing}{\spaceskip=0pt\relax}
\providecommand{\BIBentryALTinterwordstretchfactor}{4}
\providecommand{\BIBentryALTinterwordspacing}{\spaceskip=\fontdimen2\font plus
\BIBentryALTinterwordstretchfactor\fontdimen3\font minus
  \fontdimen4\font\relax}
\providecommand{\BIBforeignlanguage}[2]{{%
\expandafter\ifx\csname l@#1\endcsname\relax
\typeout{** WARNING: IEEEtran.bst: No hyphenation pattern has been}%
\typeout{** loaded for the language `#1'. Using the pattern for}%
\typeout{** the default language instead.}%
\else
\language=\csname l@#1\endcsname
\fi
#2}}
\providecommand{\BIBdecl}{\relax}
\BIBdecl

\bibitem{karaman2010optimal}
S.~Karaman and E.~Frazzoli, ``Optimal kinodynamic motion planning using
  incremental sampling-based methods,'' in \emph{49th IEEE conference on
  decision and control (CDC)}.\hskip 1em plus 0.5em minus 0.4em\relax IEEE,
  2010, pp. 7681--7687.

\bibitem{gammell2015batch}
J.~D. Gammell, S.~S. Srinivasa, and T.~D. Barfoot, ``Batch informed trees
  ({BIT}*): Sampling-based optimal planning via the heuristically guided search
  of implicit random geometric graphs,'' in \emph{2015 IEEE international
  conference on robotics and automation (ICRA)}.\hskip 1em plus 0.5em minus
  0.4em\relax IEEE, 2015, pp. 3067--3074.

\bibitem{janson2015fast}
L.~Janson, E.~Schmerling, A.~Clark, and M.~Pavone, ``Fast marching tree: A fast
  marching sampling-based method for optimal motion planning in many
  dimensions,'' \emph{The International journal of robotics research}, vol.~34,
  no.~7, pp. 883--921, 2015.

\bibitem{Lavalle2006}
S.~M. LaValle, \emph{\href{http://planning.cs.uiuc.edu/}{Planning
  Algorithms}}.\hskip 1em plus 0.5em minus 0.4em\relax Cambridge University
  Press, 2006.

\bibitem{salzman2016asymptotically}
O.~Salzman and D.~Halperin, ``Asymptotically near-optimal rrt for fast,
  high-quality motion planning,'' \emph{{IEEE} Transactions on Robotics},
  vol.~32, no.~3, pp. 473--483, 2016.

\bibitem{geraerts2007creating}
R.~Geraerts and M.~H. Overmars, ``Creating high-quality paths for motion
  planning,'' \emph{The international journal of robotics research}, vol.~26,
  no.~8, pp. 845--863, 2007.

\bibitem{toussaint2014newton}
M.~Toussaint, ``Newton methods for k-order markov constrained motion
  problems,'' \emph{arXiv preprint arXiv:1407.0414}, 2014.

\bibitem{ratliff2009chomp}
N.~Ratliff, M.~Zucker, J.~A. Bagnell, and S.~Srinivasa, ``Chomp: Gradient
  optimization techniques for efficient motion planning,'' in \emph{2009 IEEE
  International Conference on Robotics and Automation}.\hskip 1em plus 0.5em
  minus 0.4em\relax IEEE, 2009, pp. 489--494.

\bibitem{kalakrishnan2011stomp}
M.~Kalakrishnan, S.~Chitta, E.~Theodorou, P.~Pastor, and S.~Schaal, ``Stomp:
  Stochastic trajectory optimization for motion planning,'' in \emph{2011 IEEE
  international conference on robotics and automation}.\hskip 1em plus 0.5em
  minus 0.4em\relax IEEE, 2011, pp. 4569--4574.

\bibitem{schulman2014motion}
J.~Schulman, Y.~Duan, J.~Ho, A.~Lee, I.~Awwal, H.~Bradlow, J.~Pan, S.~Patil,
  K.~Goldberg, and P.~Abbeel, ``Motion planning with sequential convex
  optimization and convex collision checking,'' \emph{The International Journal
  of Robotics Research}, vol.~33, no.~9, pp. 1251--1270, 2014.

\bibitem{merkt2018leveraging}
W.~Merkt, V.~Ivan, and S.~Vijayakumar, ``Leveraging precomputation with problem
  encoding for warm-starting trajectory optimization in complex environments,''
  in \emph{IEEE International Conference on Intelligent Robots and
  Systems}.\hskip 1em plus 0.5em minus 0.4em\relax IEEE, 2018, pp. 5877--5884.

\bibitem{lembono2020memory}
T.~S. Lembono, A.~Paolillo, E.~Pignat, and S.~Calinon, ``Memory of motion for
  warm-starting trajectory optimization,'' \emph{IEEE Robotics and Automation
  Letters}, vol.~5, no.~2, pp. 2594--2601, 2020.

\bibitem{ichnowski2020deep}
J.~Ichnowski, Y.~Avigal, V.~Satish, and K.~Goldberg, ``Deep learning can
  accelerate grasp-optimized motion planning,'' \emph{Science Robotics},
  vol.~5, no.~48, p. eabd7710, 2020.

\bibitem{liu2017planning}
S.~Liu, M.~Watterson, K.~Mohta, K.~Sun, S.~Bhattacharya, C.~J. Taylor, and
  V.~Kumar, ``Planning dynamically feasible trajectories for quadrotors using
  safe flight corridors in 3-d complex environments,'' \emph{IEEE Robotics and
  Automation Letters}, vol.~2, no.~3, pp. 1688--1695, 2017.

\bibitem{toussaint2015IJCAI}
M.~Toussaint, ``Logic-geometric programming: An optimization-based approach to
  combined task and motion planning,'' in \emph{International Joint Conference
  on Artificial Intelligence}, 2015.

\bibitem{hauser2015lazy}
K.~Hauser, ``Lazy collision checking in asymptotically-optimal motion
  planning,'' in \emph{IEEE International Conference on Robotics and
  Automation}.\hskip 1em plus 0.5em minus 0.4em\relax IEEE, 2015, pp.
  2951--2957.

\bibitem{karaman2011sampling}
S.~Karaman and E.~Frazzoli, ``Sampling-based algorithms for optimal motion
  planning,'' \emph{International Journal of Robotics Research}, vol.~30,
  no.~7, pp. 846--894, 2011.

\bibitem{gammell2021asymptotically}
J.~D. Gammell and M.~P. Strub, ``Asymptotically optimal sampling-based motion
  planning methods,'' \emph{Annual Review of Control, Robotics, and Autonomous
  Systems}, vol.~4, pp. 295--318, 2021.

\bibitem{strub2021ait}
M.~P. Strub and J.~D. Gammell, ``{AIT}* and {EIT}*: Asymmetric bidirectional
  sampling-based path planning,'' \emph{arXiv preprint arXiv:2111.01877}, 2021.

\bibitem{mukadam2018continuous}
M.~Mukadam, J.~Dong, X.~Yan, F.~Dellaert, and B.~Boots, ``Continuous-time
  gaussian process motion planning via probabilistic inference,''
  \emph{International Journal of Robotics Research}, vol.~37, no.~11, pp.
  1319--1340, 2018.

\bibitem{hartmann2020planning}
V.~N. Hartmann, O.~S. Oguz, and M.~Toussaint, ``Planning planning: The path
  planner as a finite state machine,'' 2020.

\bibitem{kavraki1996probabilistic}
L.~E. Kavraki, P.~Svestka, J.-C. Latombe, and M.~H. Overmars, ``Probabilistic
  roadmaps for path planning in high-dimensional configuration spaces,''
  \emph{IEEE transactions on Robotics and Automation}, vol.~12, no.~4, pp.
  566--580, 1996.

\bibitem{kim2003extracting}
J.~Kim, R.~A. Pearce, and N.~M. Amato, ``Extracting optimal paths from roadmaps
  for motion planning,'' in \emph{IEEE International Conference on Robotics and
  Automation}, vol.~2.\hskip 1em plus 0.5em minus 0.4em\relax IEEE, 2003, pp.
  2424--2429.

\bibitem{choudhury2016regionally}
S.~Choudhury, J.~D. Gammell, T.~D. Barfoot, S.~S. Srinivasa, and S.~Scherer,
  ``Regionally accelerated batch informed trees ({rabit}*): A framework to
  integrate local information into optimal path planning,'' in \emph{2016 IEEE
  International Conference on Robotics and Automation (ICRA)}.\hskip 1em plus
  0.5em minus 0.4em\relax IEEE, 2016, pp. 4207--4214.

\bibitem{alwala2020joint}
K.~V. Alwala and M.~Mukadam, ``Joint sampling and trajectory optimization over
  graphs for online motion planning,'' in \emph{IEEE International Conference
  on Intelligent Robots and Systems}.\hskip 1em plus 0.5em minus 0.4em\relax
  IEEE, 2020, pp. 4700--4707.

\bibitem{kuntz2020fast}
A.~Kuntz, C.~Bowen, and R.~Alterovitz, ``Fast anytime motion planning in point
  clouds by interleaving sampling and interior point optimization,'' in
  \emph{Robotics Research}.\hskip 1em plus 0.5em minus 0.4em\relax Springer,
  2020, pp. 929--945.

\bibitem{xanthidis2020navigation}
M.~Xanthidis, N.~Karapetyan, H.~Damron, S.~Rahman, J.~Johnson, A.~O’Connell,
  J.~M. O’Kane, and I.~Rekleitis, ``Navigation in the presence of obstacles
  for an agile autonomous underwater vehicle,'' in \emph{2020 IEEE
  International Conference on Robotics and Automation (ICRA)}.\hskip 1em plus
  0.5em minus 0.4em\relax IEEE, 2020, pp. 892--899.

\bibitem{Orthey2020RAL}
A.~Orthey, B.~Fr\'{e}sz, and M.~Toussaint, ``Motion planning explorer:
  Visualizing local minima using a local-minima tree,'' \emph{IEEE Robotics and
  Automation Letters}, vol.~5, no.~2, pp. 346--353, April 2020.

\bibitem{dai2018improving}
S.~Dai, M.~Orton, S.~Schaffert, A.~Hofmann, and B.~Williams, ``Improving
  trajectory optimization using a roadmap framework,'' in \emph{2018 IEEE/RSJ
  International Conference on Intelligent Robots and Systems (IROS)}.\hskip 1em
  plus 0.5em minus 0.4em\relax IEEE, 2018, pp. 8674--8681.

\bibitem{park2015parallel}
C.~Park, F.~Rabe, S.~Sharma, C.~Scheurer, U.~E. Zimmermann, and D.~Manocha,
  ``Parallel cartesian planning in dynamic environments using constrained
  trajectory planning,'' in \emph{2015 IEEE-RAS 15th International Conference
  on Humanoid Robots (Humanoids)}.\hskip 1em plus 0.5em minus 0.4em\relax IEEE,
  2015, pp. 983--990.

\bibitem{Orthey2020WAFR}
A.~Orthey and M.~Toussaint, ``Visualizing local minima in multi-robot motion
  planning using multilevel morse theory,'' in \emph{Workshop on the
  Algorithmic Foundations of Robotics}, 2020.

\bibitem{penrose2003random}
M.~Penrose, \emph{Random geometric graphs}.\hskip 1em plus 0.5em minus
  0.4em\relax OUP Oxford, 2003, vol.~5.

\bibitem{gammell2020batch}
J.~D. Gammell, T.~D. Barfoot, and S.~S. Srinivasa, ``Batch informed trees
  (bit*): Informed asymptotically optimal anytime search,'' \emph{The
  International Journal of Robotics Research}, vol.~39, no.~5, pp. 543--567,
  2020.

\bibitem{sucan2012open}
I.~A. Sucan, M.~Moll, and L.~E. Kavraki, ``The open motion planning library,''
  \emph{IEEE Robotics \& Automation Magazine}, vol.~19, no.~4, pp. 72--82,
  2012.

\end{thebibliography}
}
\end{document}